# Inadequate contrast ratio of road markings as an indicator for ADAS failure

Novel Certad [a], Cristina Olaverri-Monreal [a], Friedrich Wiesinger [b], Tomasz E. Burghardt *[b]

[a] *Johannes Kepler University Linz; Department Intelligent Transport Systems, Altenberger Straße 66C, 4040 Linz, Austria*
[b] *M. Swarovski GmbH; Wipark, 14 Straße 11, 3363 Neufurth, Austria*
* Corresponding author. Email address: tomasz.burghardt@swarco.com

**Abstract**

Road markings were reported as critical road safety features, equally needed for both human drivers and for machine vision technologies utilised by advanced driver assistance systems (ADAS) and in driving automation. Visibility of road markings is achieved because of their colour contrasting with the roadway surface. During recent testing of an open-source camera-based ADAS under several visibility conditions (day, night, rain, glare), significant failures in trajectory planning were recorded and quantified. Consistently, better ADAS reliability under poor visibility conditions was achieved with Type II road markings (i.e. structured markings, facilitating moisture drainage) as compared to Type I road marking (i.e. flat lines). To further understand these failures, analysis of contrast ratio of road markings, which the tested ADAS was detecting for traffic lane recognition, was performed.

The highest contrast ratio (>0.5, calculated per Michelson equation) was measured at night in the absence of confounding factors, with statistically significant difference of 0.1 in favour of Type II road markings over Type I. Under daylight conditions, contrast ratio was reduced, with slightly higher values measured with Type I. The presence of rain or wet roads caused the deterioration of the contrast ratio, with Type II road markings exhibiting significantly higher contrast ratio than Type I, even though the values were low (<0.1).

These findings matched the output of the ADAS related to traffic lane detection and underlined the importance of road marking's visibility. Inadequate lane recognition by ADAS was associated with very low contrast ratio of road markings indeed. Importantly, specific minimum contrast ratio value could not be found, which was due to the complexity of ADAS algorithms that rely simultaneously on edge detection and recognition of other roadway features. The results also indicate that while the issue of blinding by glare cannot be solved at present with camera-based ADAS input, the visibility decrease due to rain could be meaningfully alleviated by the use of Type II structured road markings. Hence, such road markings should be installed to increase resiliency and dependability of road network, to improve reliability of ADAS, and foremost to augment comfort and safety of human drivers.

## 1. Introduction

The use of road markings (RM) as means of improving road traffic safety reportedly dates back to 1912. Since then, they are commonly used to delimit travel paths for users of almost all roads. The utilisation of RM was calculated to bring enormous financial benefits due to decreasing of the number and severity of accidents through channelling and organising traffic [1]. RM are equally needed for human drivers and for ADAS that are nowadays implemented in almost all new vehicles [2,3]. Amongst advantages of RM one should list established and dependable material and application technologies, reliability, equal usefulness for both human drivers and for ADAS, their existing presence on roads, and efficiency without external energy supply. Hence, RM cannot be replaced within any currently known technologies.

There is a plethora of research on the effects of RM on driver behaviour [4]. Similarly, numerous articles were written on the recognition of RM by ADAS [5]. However, such research appeared to be missing the effect of materials used for RM on their visibility – in a series of articles this issue was recently addressed by us: the profound importance of materials selection on the achieved visibility of RM under various conditions was examined under both laboratory and field conditions [6,7,8]. It was repeatedly shown that RM of Type II (structured, capable of improved moisture drainage) were outperforming RM of Type I (flat lines, not designed for enhanced visibility under conditions of wetness). To understand these differences, one should observe the intricacies of the technology of RM. As materials, RM belong to speciality heavy duty industrial coatings, highly unique because of comprising two distinct layers that must cooperate to form the final functional product: the bottom colour layer and the top retroreflective layer [9].

For correct functioning, RM must be visible to drivers and to ADAS, which is possible because of a contrast between the roadway and the markings [10]. Such contrast can be quantified through contrast ratio (CR). Even though there were attempts to establish a minimum CR for proper functioning of ADAS, with claims that Weber contrast of >2.0 would be needed [11], the issue does not seem to be so easy to quantify and numerous doubts remain. Research indicated that CR value itself was not as important as edge recognition [12]. In addition, the use of Weber equation for CR was deemed inadequate due to the possibility of infinite values; instead, the use of Michelson contrast, with the range truncated at 1.0, appeared to be better representing the perception. Review of the prior literature on this topic, due to reported inconsistencies that would demand excessive explanations, remains beyond the scope of this report.

Our recent effort toward better understanding of the behaviour of ADAS in response to RM of Type I and Type II included evaluation of an open-source software under various visibility conditions, with quantification of failures in trajectory planning [13]. As an expansion of that analyses, CR of the RM that were used by the software for trajectory planning was evaluated. The purpose of the work described herein was checking whether low CR could be an indicator of impending ADAS failures. The described outcome would be useful not only for software engineers working on improvement of the algorithms for automated driving features, but foremost for the road administrators who should prepare the roads for the implementation of ADAS and ought to strive to increase road safety through augmenting visibility of RM.

## 2. Methodology

The evaluation was done at a closed course proving grounds in St. Valentin, Austria. The proving grounds are equipped with a rain simulator and a high intensity lamp (delivering luminous flux 7211 lm in a narrow spectral distribution) as a glare source. Traffic lanes (net width 3.0–3.2 m) at the test field were delimited with RM of Type I (flat lines, low retroreflectivity, poor visibility during rain) and of Type II (structured materials, high retroreflectivity, good visibility during rain), applied commercially as longitudinal lines according to the typical motorway arrangement in Austria (lines 12 cm wide, 6 m long, spaced 12 m apart). The analysis described herein was done on RM located in two regions: before the test vehicle entered the poor visibility region under the rain simulator ('control' region, with dry road surface) and under the rain simulator ('test conditions' region). Further details about the materials, equipment, and test set-up were described elsewhere [13].

One of the outputs of the evaluated ADAS software was a 20 Hz video recording (256×512 pixels). The RM visible in representative images were analysed for CR using Michelson equation (eq. 1), where $CR_M$ – Michelson contrast ratio, $L_m$ – luminance of the marked area (average from 9 pixels), and $L_b$ – luminance of the background (average from 9 pixels neighbouring the pixels used for $L_m$). For each analysed RM stripe, CR was sampled over at least four (and up to ten) locations. The selection of Michelson equation ($CR_M$) was done because it is providing information about data quality (signal to noise level), which seems more appropriate than the use of Weber equation that is more suitable to analyses related to lighting. Possible Michelson CR values are between −1 and +1 (in case of Weber contrast, the values between −1 and +∞ are possible). In cases of negative CR, absolute values were taken for the purpose of this analysis, with an assumption that they would still furnish an edge and delimit a travel path that could be detected by the software.

$$CR_M = \frac{L_m - L_b}{L_m + L_b} \qquad (1)$$

The conditions for analyses are described in Table 1. In most of the cases, visual identification of the RM did not extend past the first stripe (in some cases also beginning of the second stripe) after entering the 'test conditions' region under the rain simulator. This inadequacy of the delivered images could be solved through employment of a separate camcorder with different dynamic range; however, since it would imply testing of different input than was used by the software.

Table 1. Conditions for testing.

| Conditions code | Lighting[a] | Visibility[b] | Road surface[b] | Average travel speed [km/h] | Simulated rainfall [mm/h][b] |
|---|---|---|---|---|---|
| ND-30-00 | Night | Clear | Dry | 30 | – |
| DG-08-00 | Dusk | Glare | Dry | 8 | – |
| DW-30-35 | Day | Wet surface | Wet | 30 | 35 |
| DG-08-35 | Dusk | Glare | Wet | 8 | 35 |
| NG-08-00 | Night | Glare | Wet | 8 | – |

[a]Applies to both 'control' and 'test conditions' regions. [b]Applies to 'test conditions' region.

## 3. Results

The outcome from the analyses that includes the CR in both regions and the statistical significance (Student's t-test, one-tailed distribution, heteroscedastic, 95% confidence interval) of the differences between the two tested types of RM is shown in Table 2 and summary of the results is charted in Figure 1. Data given in Table 2 is supplemented with synopsis of previously reported detection probabilities of RM [13]. Out of five visibility conditions, only in the presence of glare and during daytime, RM of Type I were providing higher CR than Type II. Most likely this was due to the three-dimensional structure of Type II RM, which resulted in shadowing and lower effective surface coverage, but proper elucidation of this phenomenon is noted hereby as a research need. Noteworthy was the lack of visibility of the second stripe (i.e. distance >18 m) under some conditions. While analysing the results for the conditions of wetness, one should keep in mind that the issues associated with rendering of wet dark surfaces, which appear darker and thus artificially increase CR even though visibility decreases [14], were not addressed herein; nonetheless, the effect did not seem to have affected the reported outcome.

Table 2. Contrast ratio of road markings the two analysed regions.

| Conditions code | Region | Michelson contrast ratio ($CR_M$) | | T-test (Type I / Type II) | Second stripe visibility | $CR_M$ result[a] | Detection probability[b] | |
|---|---|---|---|---|---|---|---|---|
| | | Type I | Type II | | | | Type I | Type II |
| ND-30-00 | Control | 0.55 | 0.65 | 0.043 | Visible | I < II | 0.83 | 0.73 |
| | Test conditions | 0.59 | 0.66 | 0.007 | Visible | I < II | 0.87 | 0.84 |
| DG-08-00 | Control | 0.23 | 0.21 | 0.221 | Visible | I = II | 0.91 | 0.85 |
| | Test conditions | 0.24 | 0.18 | 0.013 | Visible | I ≫ II | 0.81 | 0.77 |
| DW-30-35 | Control | 0.25 | 0.19 | 0.000 | Visible | I ≫ II | 0.88 | 0.76 |
| | Test conditions | 0.06 | 0.09 | 0.008 | Not visible | I ≪ II | 0.71 | 0.81 |
| DG-08-35 | Control | 0.31 | 0.45 | 0.002 | Visible | I ≪ II | 0.84 | 0.89 |
| | Test conditions | 0.03 | 0.02 | 0.174[c] | Not visible | I = II | 0.53 | 0.39 |
| NG-08-00 | Control | 0.24 | 0.45 | 0.003 | Visible | I ≪ II | 0.86 | 0.95 |
| | Test conditions | 0.02 | 0.08 | 0.000 | Not visible | I ≪ II | 0.56 | 0.87 |

[a]Comparison in CR performance of Type I and Type II RM. [b]Detection probability of RM by the tested ADAS software [13]. [c]Difference not statistically significant.

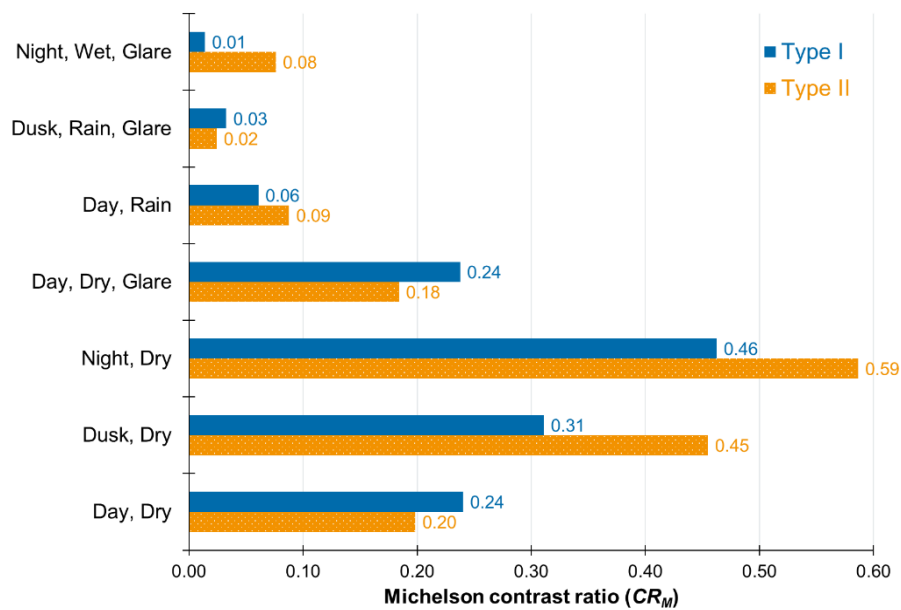

Figure 1. Contrast ratio measured under various conditions.

To answer the question whether failures of the tested ADAS software could be predicted through the measurements of CR, Pearson correlation was calculated. In case of the 'control' regions, the correlations were negative, −0.28 for Type I and −0.68 for Type II. Contrariwise, in the 'test conditions' regions the correlations were positive and quite strong: 0.85 for Type I and 0.42 for Type II. Hence, it could be concluded that, under poor visibility conditions, performance of the evaluated ADAS software was strongly dependent on the visibility of RM.

## 4. Conclusions

The results demonstrated for the nth time that RM of Type II are capable of delivering better visibility, assessed through CR of representative images. As such, structured RM are more suitable for both human drivers and for camera-based ADAS software than flat line Type I RM. Major failures of traffic lane recognition by the evaluated ADAS were measured during inclement weather conditions and herein it was shown that poor CR of RM was very likely a major contributor to those failures.

Lane keeping assistant (LKA), which strongly depends on the recognition of RM [15], was recently made mandatory in all new vehicles sold in European Union, which makes the outcome of this study of utmost practical importance. Since the consumers are being forced to purchase vehicles capable of RM recognition, it would be only reasonable to expect that the infrastructure read by the LKA would be maintained appropriately, so the feature could be effectively and reliably utilised. Therefore, it is highly surprising that RM are so often permitted to be grossly neglected. The use of Type II RM that provide the best visibility for human drivers, for the LKA, and other ADAS should therefore be made mandatory and all RM ought to be maintained properly with renewals done upon the decrease of retroreflectivity to below 150 mcd/m²/lx in dry and 35 mcd/m²/lx in wet conditions, as was recommended by the European Union Road Federation [16]. Simultaneously, better visibility of Type II RM under poor conditions would benefit human drivers, as was repeatedly reported [17,18], which could be correlated with the decrease in accidents [19].

The following general conclusions can be drawn about the effects of various visibility conditions on CR of RM of Type I and Type II:

- Lighting. With a good level of natural light, RM of Type I provided CR higher than or equal to RM of Type II. Reducing the external light level increased the CR in favour of Type II.
- Dry conditions at daytime: RM of Type I were advantageous.
- Wet conditions at night time: RM of Type II were clearly advantageous.
- Rain. The presence of precipitation or wet roadway caused very meaningful decrease in CR for both types of RM, but Type II was clearly advantageous.
- Glare. Difficult to analyse and inconclusive results due to profound effect of the glare source position. This effect is noted herein as an important future research need.